\pdfoutput=1

\documentclass[11pt]{article}

\usepackage[final]{acl}

\usepackage{times}
\usepackage{latexsym}
\usepackage{tikz}
\usepackage{circledsteps}
\usepackage[T1]{fontenc} 
\usepackage[utf8]{inputenc} 
\usepackage{microtype} 
\usepackage{inconsolata} 
\usepackage{graphicx} 

\usepackage{CJKutf8}

\usepackage{diagbox}
\usepackage{makecell}
\usepackage{comment}
\usepackage{pgfplots}
\usepackage{amsmath, xcolor}  
\definecolor{lightblue}{RGB}{173, 216, 230}  
\pgfplotsset{compat=1.18}

\title{BreezyVoice: Adapting TTS for Taiwanese Mandarin with Enhanced Polyphone Disambiguation -- Challenges and Insights}

\author{
    Chan-Jan Hsu$^{1}$\thanks{These authors contributed equally.}, Yi-Cheng Lin$^{2*}$, Chia-Chun Lin$^{1}$, Wei-Chih Chen$^{2}$, Ho Lam Chung$^{2}$, \\
    {\bf Chen-An Li$^{2}$, Yi-Chang Chen$^{1}$, Chien-Yu Yu$^{1}$, Ming-Ji Lee$^{1}$}, \\
    {\bf Chien-Cheng Chen$^{2}$, Ru-Heng Huang$^{1}$, Hung-yi Lee$^{2}$, Da-Shan Shiu$^{1}$} \\
    $^{1}$MediaTek Research \quad\quad $^{2}$National Taiwan University
}

\begin{document}
\maketitle

\begin{abstract}
We present BreezyVoice, a Text-to-Speech (TTS) system specifically adapted for Taiwanese Mandarin, highlighting phonetic control abilities to address the unique challenges of polyphone disambiguation in the language. Building upon CosyVoice, we incorporate a $S^{3}$ tokenizer, a large language model (LLM), an optimal-transport conditional flow matching model  (OT-CFM), and a grapheme to phoneme prediction model, to generate realistic speech that closely mimics human utterances. Our evaluation demonstrates BreezyVoice's superior performance in both general and code-switching contexts, highlighting its robustness and effectiveness in generating high-fidelity speech. Additionally, we address the challenges of generalizability in modeling long-tail speakers and polyphone disambiguation. Our approach significantly enhances performance and offers valuable insights into the workings of neural codec TTS systems.
\end{abstract}

\section{Introduction}

Text-to-Speech (TTS) technology has been widely adopted in customer service and accessible tools. Recently, the significant advancements in virtual assistants \cite{hurst2024gpt} increased the frequency of breadth of TTS utilization, which has prompted a shift away from the mechanical voice outputs towards speech that closely mimics the nuances of human conversation. Aside from immediate practical scenarios, we also highlight the emerging use case of synthetic data generation, as it offers extensive coverage for spontaneous speech in controllable written domains. The benefits of using these otherwise inaccessible synthetic data for training purposes have been well-documented in previous literature \cite{lin2022analyzing, yang2024building}.

In response to the increasing expectations for high-quality outcomes, it is particularly advantageous to address localized TTS needs, from adaptation of an existing system, leveraging the knowledge transfer potential from  pretrained large-scale corpora. Recent realistic TTS systems have reported training data of approximately 170,000 hours \cite{du2024cosyvoice}, which is comparable to the scale of training large ASR systems \cite{radford2023robust}, where adaptation efforts for specialization are common \cite{tseng2024leave}.

In this work, we build upon CosyVoice to develop a Taiwanese Mandarin TTS system, which we name as BreezyVoice. BreezyVoice demonstrates superior performance compared to commercial TTS systems, both in general and code-switching contexts. Beyond the training process, we tackle generalizability challenges in the adapted system from lower-resourced training, primarily involving the correct modeling of long-tail speaker characteristics and the polyphonic characters inherent in the Taiwanese Mandarin language. Our detailed analysis provide valuable insights into the workings
of neural codec TTS systems.

\section{Background}
\subsection{Unit-TTS Systems}
Advanced speech compression codec systems \cite{defossez2022high, du2024funcodec} have successfully encapsulated verbose speech information into quantized speech units. These units are seamlessly adopted for pursuing Text-to-Speech (TTS) systems \cite{ju2024naturalspeech}, via incorporating a module that predicts the compressed units from text along with an optional speaker embedding. The design of the text-to-unit model is highly dependent on the granularity of the units to predict. For example, multi-layered codes are often predicted recursively using a non-autoregressive model \cite{wang2023neural, chen2024vall}, whereas single-layered codes are structured as a sequence-to-sequence generation task \cite{du2024cosyvoice}. From this perspective, we identified the benefits of single-layered supervised semantic tokens from Coisyvoice, which dynamically encode essential content and prosody information with a high compression rate.

\subsection{The Necessity of Taiwanese Mandarin Conversion}

The default language of the Cosyvoice model is Mandarin, which is mutually intelligible with Taiwanese Mandarin, but also features various lexical and phonological differences \cite{bradley1992chinese}. Further, Taiwanese Mandarin employs Traditional Chinese characters as the surface form, which significantly overlap with Cantonese in written form but differ greatly in pronunciation, and there have been scholars that regard them as separate languages \cite{mair1991chinese}. As the CosyVoice model predominantly associates Traditional Chinese characters with Cantonese, it is crucial to ensure that the model is adequately trained with Taiwanese Mandarin data to guarantee performance on this domain.

\subsection{Polyphone Disambiguation and Phonetic Control for Boosting TTS Accuracy}
\begin{CJK*}{UTF8}{bsmi}
Polyphone disambiguation is crucial for Mandarin grapheme-to-phoneme (G2P) conversion, essential for accurate text-to-speech (TTS) systems. Mandarin characters can have multiple pronunciations based on context\footnote{For instance, the character ``行'' has different pronunciations: ``ㄒㄧㄥˊ'' (to walk) or ``ㄏㄤˊ'' (a row).}, making disambiguation vital, as incorrect pronunciations make phrases unintelligible. To alleviate potential errors in a speech synthesis system, the phonetic controllability of system is highly desired, as it allows us to preamptively augment the accurate Mandarin Phonetic Symbols predicted from a Neural Model g2pW \cite{yc_g2pw}. We perform phonteic augmenetations using g2pW in both training and inference to boost TTS accuracy.
\end{CJK*}

\section{Methodology}

\subsection{BreezyVoice Architecture}  

The BreezyVoice Framework consists of four components: A Supervised Semantic Speech (S3) Tokenizer, a large language model (LLM), an optimal-transport conditional flow matching model (OT-CFM) \cite{mehta2024matcha}, and a grapheme to phoneme
prediction model g2pW. The following sections describe the components in detail.

\subsubsection{Supervised Semantic Speech Tokenizer}
The speech tokenizer is a Speech Encoder with a vector quantization layer \cite{zeghidour2021soundstream} to output discrete units. The semantic supervision is derived from training these units on a supervised Automatic Speech Recognition (ASR) task. In the one-shot speech cloning setting, the speech tokenizer is used to generate units of the conditioning speaker sample.
\subsubsection{Large Language Model for TTS}

In our framework, the text-to-unit generation task is fulfilled with a large language model (LLM). The model takes tokenized text data and a speaker embedding as inputs, and outputs speech units.


During training, the LLM uses teacher forcing technique, taking advantage of the decoding structure to  minimizing the cross-entropy losses of the predicted speech sequences. At inference, the LLM generates speech tokens in an auto-regressive manner, providing a foundation for high-fidelity speech synthesis.

\subsubsection{Optimal-transport Conditional Flow Matching}  

In our framework, the transformation of speech tokens into a Mel spectrogram is achieved using an optimal-transport conditional flow matching (OT-CFM) model. OT-CFM models leverage
optimal transport (OT) to learn a dynamic gradual transformation, iteratively refining a prior into the meaningful data distribution. In particular, our OT-CFM model generates continuous spectrograms that accurately reflect the time-frequency structure of an utterance, from conditioning on speaker characteristics. The input to this stage includes LLM-generated speech tokens, a speaker embedding, and a masked Mel spectrogram where the masked part is to be reconstructed.

\subsubsection{Phoneme Prediction Model}



In this work, we utilize g2pW  as our phoneme prediction model. g2pW is an open-source model that currently holds the highest performance on the CPP benchmark \cite{park2020g2pm}. This model employs learnable softmax weights to condition the outputs of BERT \cite{kenton2019bert}. The authors demonstrate that using a conditional weighted softmax, which is conditioned on the polyphonic character of interest and its part-of-speech (POS) tagging, enhances the performance of polyphone disambiguation.

\subsection{Implementation Details}

We now describe implementation details of our approach, covering data preparation and training.

\subsubsection{Data Preparation}
We use a diverse collection of publicly available speech data and utilize its paired transcriptions. When transcriptions are unavailable, we use Generative Fusion Decoding \cite{hsu2024let}, an LLM-assisted automatic speech recognition system, to conduct pseudo-labeling.

\begin{figure}[ht]
    \centering
    \includegraphics[width=\linewidth]{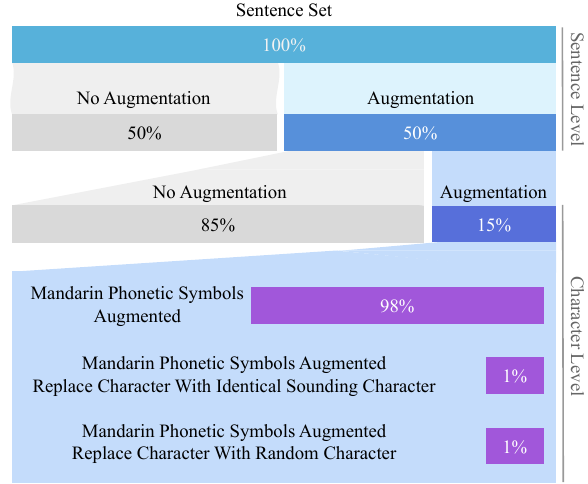}
    \caption{The augmentation pipeline for integrating Mandarin phonetic symbols employs a decision tree at both the sentence and character levels to diversify the inputs. Additionally, we partially add noise to some of the inputs to prioritize the model's attentiveness to the phonetic symbols.}
    \label{fig:aug-and-noise}
\end{figure}

To ensure that punctuations are present in the appropriate places, we use the Breeze-7B LLM \cite{hsu2024breeze} to augment the punctuations in the raw text. This aligns the text with the downstream use, which is likely to contain punctuations. Additionally, to make the model understand phonetic representations, we randomly augment text with mandarin phonetic symbols. The augmentation and noising logic is inspired from BERT \cite{devlin-etal-2019-bert} and shown in Figure~\ref{fig:aug-and-noise}. The emphasis of the augmentation technique is to maximize the model's efficiency in learning the pronunciation of Chinese characters, while developing the model's ability to prioritize attending to supplementary symbols when they occur.

\subsubsection{Training Details}

We train only the LLM, which has been reported to mostly capture semantic and prosody information. We train the model for three epochs, with a flat learning rate of 1e-4. We use dynamic batching with a length of 8000 to accommodate training samples that have significantly varied lengths.

\section{Experimental Settings}

Our experiments are two composed of two parts:  comparisons with other systems, and evaluations on voice cloning.

Our comprehensive survey reveals that the leading Taiwanese Mandarin TTS systems are all proprietary services. Since many online services use one or more preset voices and lack cloning features, it is impossible to compare them on that aspect. As a result, we primarily focus on comparing audio quality, by selecting an utterance from a generic female speaker that closely matches the preset voice of multiple TTS systems to serve as the speaker conditioner for our model. The evaluation targets are both in Chinese and bilingual, considering that contemporary language usage frequently incorporates foreign terms, especially in contexts such as entity names, technical jargon, and cultural references.

For voice cloning evaluations, we evaluate on PER and speaker similarity using samples generated from a diverse range of speakers to gauge the robustness of our system.

\subsection{Evaluation Dataset}

The dataset consists of a multi-speaker corpus for speech conditioning, and a textual corpus for generation.  For the multi-speaker corpus, we have collected 15 utterances sampled from the FormosaSpeech \cite{liao2020formosa} corpus, and 100 utterances of spontaneous recordings in an environment with limited background noise. The speech samples come from a diverse group of Taiwanese Mandarin speakers varying in age, gender, pitch, and other characteristics. Each utterance ranges from 5 to 15 seconds in duration, with their spoken word targets manually annotated and verified.

The \textbf{Traditional Chinese Monologue Dataset (TCMD)} is derived from a Traditional Chinese conversational dataset and includes exclamations, questions, and regular sentences, which naturally diversify the intended expressions without relying on explicit expression annotations. The \textbf{Traditional Chinese Code-switching Dataset (TCCSD)} is a code-switching text corpus, and is generated and categorized into five categories from scouring read-world data for natural context-switching scenarios. The categories are: general words, entities, abbreviations, Taiwan-related toponyms, and full sentences. Phonetic accuracy and prosodic appropriateness are the primary evaluation criteria. Both datasets were collected independently, without prior knowledge of BreezyVoice or other TTS systems.
\subsection{Inference Strategies}
The standard inference pipeline of BreezyVoice is derived from CosyVoice \cite{du2024cosyvoice}. Highlighting the input and output data types, the pipeline is described as such:
\begin{equation}
    \textbf{U}_{output} = LLM(v, \textbf{Y}_{cond}, \textbf{Y}_{output}, \textbf{U}_{cond})
\end{equation}
\begin{equation}
    \textbf{X}_{output} = CFM(v, \textbf{U}_{cond}, \textbf{U}_{output}, \textbf{X}_{cond})
\end{equation}
where $v$ denotes the speech representation, $CFM$ denotes the OT-CFM model, $\textbf{Y}$ denotes the byte-pair encoded text, $\textbf{U}$ denotes the derived speech unit, and $\textbf{X}$ denotes the mel-spectrogram. The subscripts ``cond'' and ``output'' denote the sources of these elements, implying from conditioning speech or from speech to be generated, respectively. The conversion from mel-spectrogram to audio is completed with an static algorithmic approach and excluded in the discussion. Additionally, there is the option to augment Mandarin phonetic symbols.
\begin{equation}
    \textbf{Y}_{augmented} = \text{g2pW}(\textbf{Y})
\end{equation}

The multitude of conditioners, coupled with dropout training of previous work \cite{du2024cosyvoice}, prompts an exploration of alternative inference possibilities. 
\textit{What is the optimal inference strategy for various scenarios, and when are augmentation necessary?} 
These insights into the effects of conditioning and auxiliary inputs are crucial for making informed design decisions in the industrial deployment of the system, which we examine in Section~\ref{sec:analysis}.
\section{Experimental Results}

The following section describes our experimental results on BreezyVoice, using the standard inference pipeline to generate speech audio from a given speaker and specified text. 

\subsection{Comperlative Evaluations with Competing Systems}
To allow for a head-to-head comparison, an iconic speaker is isolated from FormosaSpeech to serve as the reference anchor. This selection is made purely based on the similarity to the preset voice of competing systems. 

In subjective evaluations, the annotators were tasked to evaluate the audio sample based on authenticity and audio quality (Appendix~\ref{sec:appendix_annotations}) The outputs were ranked from most to least favorable based on their overall performance on these two combined metrics (with ties allowed), using a blind test interface. Three annotators were given equal pay to participate in this test, and each assigned preferences for five different systems on 10 samples of TCMD, resulting in a total of 30 comparative evaluations between any two models.

\begin{figure}
\begin{tikzpicture}
\begin{axis}[
    width=7.0cm,
    height=5cm,
    xbar stacked,
    bar width=0.6cm,
    symbolic y coords={Service M, Service U, Service Y, Service Z},
    ytick=data,
    xmin=0, xmax=100,
    xtick={0,25,50,75,100},
    xticklabel={\pgfmathprintnumber{\tick}\%},
    xlabel=Win Rate,
    legend style={at={(0.4,1.1)}, anchor=south, legend columns=-1},
    legend cell align={left},
    nodes near coords,
    point meta=explicit,  
    every node near coord/.append style={font=\small},
    enlarge y limits=0.2
]

\addplot[fill={rgb,255:red,245;green,154;blue,0}]
coordinates {
    (100,Service M)[30]
    (73.33,Service U)[22]
    (63.33,Service Y)[19]
    (80,Service Z)[24]
};

\addplot[fill={rgb,255:red,255;green,194;blue,102}]
coordinates {
    (0,Service M)[0]
    (0,Service U)[0]
    (3.33,Service Y)[1]
    (0,Service Z)[0]
};

\addplot[fill={rgb,255:red,255;green,230;blue,179}]
coordinates {
    (0,Service M)[0]
    (26.67,Service U)[8]
    (33.33,Service Y)[10]
    (20,Service Z)[6]
};

\legend{BreezyVoice Wins, Tie, Loses}
\end{axis}
\end{tikzpicture}
\caption{Human preference evaluation comparing BreezyVoice with four competing systems across 30 comparisons on the TCMD dataset. The results demonstrate BreezyVoice's consistent superior performance.}
    \label{fig:human-pref}
\end{figure}
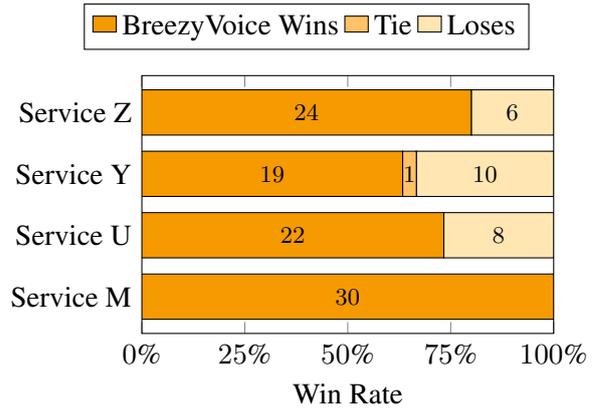

The results in Figure~\ref{fig:human-pref} demonstrate that BreezyVoice excelled in subjective assessments, consistently achieving higher win rates compared to the other four services (anonymized with letters Z, Y, U, M), which validates the model's competence in generating authentic quality samples. 

The subjective evaluations are backed up by strong objective scores of PER (Phoneme Error Rate) and SSL-MOS scores. The PER is calculated with predicted phonetic symbols from cascading Whisper-large with g2pW. The MOS predictor is an SSL-MOS trained with a Taiwanese Mandarin dataset, TMHINT-QI \cite{zezario2023study}, implemented in sheet toolkit\footnote{https://huggingface.co/unilight/sheet-models/tree/main/tmhint-qi/sslmos/2337}. The results are shown in Table~\ref{tab:objective}.

\begin{table}[h!]
  \centering
  \small
  \begin{tabular}{|l|c|c|c|c|c|}
    \hline
    \diagbox[width=2cm, height=1cm]{}{Model} & \makecell{\textbf{B} \\ \textbf{(ours)}}  & Z & Y&  U & M \\
    \hline
    \textbf{PER(\%)} & 0.8  & 1.17 & 1.45 & 0.69 & 0.43 \\
    \hline
    \textbf{SSL-MOS} & 4.46 & 4.63 & 3.84 & 3.46 & 4.40 \\
    \hline
  \end{tabular}
  \caption{Objective evaluation of BreezyVoice against four competing systems on the TCMD dataset, measured by Phoneme Error Rate (PER) and SSL-MOS scores.}
  \label{tab:objective}
\end{table}


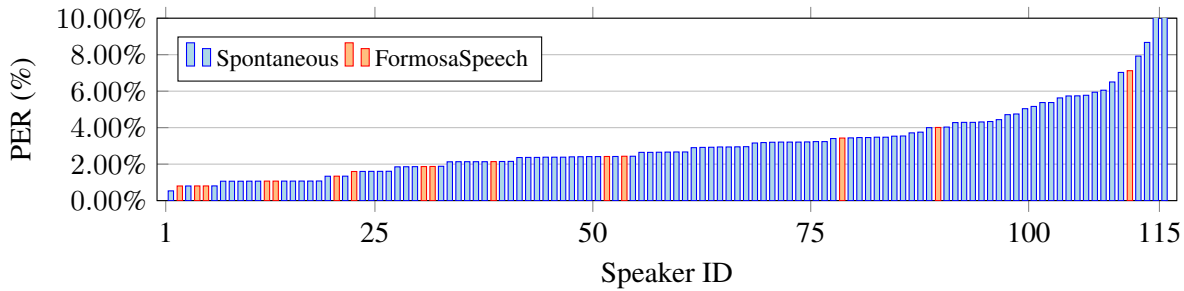
\begin{figure*}[htbp]
    \centering
    \begin{tikzpicture}[]
        \begin{axis}[
            ybar,                              
            bar width=2pt,                    
            width=15cm,
            height=4cm,                       
            xlabel={Speaker ID },
            ylabel={PER (\%)},              
            xtick={1, 25, 50, 75, 100, 115}, 
            xticklabels={1, 25, 50, 75, 100, 115},
            xmin=0, xmax=117,
            ymin=0, ymax=0.1,  
            ymajorgrids,
            yticklabel style={/pgf/number format/fixed,/pgf/number format/fixed zerofill,/pgf/number format/precision=2}, 
            legend pos=north west,            
            legend style={
                font=\small,
                legend columns=-1,            
                at={(0.02,0.9)},              
                anchor=north west
            },
            legend entries={Spontaneous, FormosaSpeech}, %
            clip=true, 
            yticklabel={\pgfmathparse{\tick*100}\pgfmathprintnumber{\pgfmathresult}\%}, 
        ]
            
            \addplot+[
            ybar,
            fill=lightblue,
        ] coordinates {
            (2.2, 0.005333333333333333) (4.2, 0.008021390374331552) (7.2, 0.008064516129032258) (8.2, 0.010638297872340425) (9.2, 0.010638297872340425) (10.2, 0.010638297872340425) (11.2, 0.010666666666666666) (12.2, 0.010666666666666666) (15.2, 0.0106951871657754) (16.2, 0.0106951871657754) (17.2, 0.010723860589812333) (18.2, 0.010723860589812333) (19.2, 0.010752688172043012) (20.2, 0.013368983957219251) (22.2, 0.013404825737265416) (24.2, 0.016) (25.2, 0.016042780748663103) (26.2, 0.016042780748663103) (27.2, 0.0160857908847185) (28.2, 0.018518518518518517) (29.2, 0.01856763925729443) (30.2, 0.018617021276595744) (33.2, 0.01881720430107527) (34.2, 0.02127659574468085) (35.2, 0.021333333333333333) (36.2, 0.021333333333333333) (37.2, 0.021333333333333333) (38.2, 0.021333333333333333) (40.2, 0.021447721179624665) (41.2, 0.021505376344086023) (42.2, 0.023622047244094488) (43.2, 0.02368421052631579) (44.2, 0.02368421052631579) (45.2, 0.023809523809523808) (46.2, 0.023809523809523808) (47.2, 0.023809523809523808) (48.2, 0.024) (49.2, 0.02406417112299465) (50.2, 0.024128686327077747) (51.2, 0.024128686327077747) (53.2, 0.024193548387096774) (55.2, 0.024324324324324326) (56.2, 0.026455026455026454) (57.2, 0.026455026455026454) (58.2, 0.026525198938992044) (59.2, 0.026595744680851064) (60.2, 0.02666666666666667) (61.2, 0.02666666666666667) (62.2, 0.029023746701846966) (63.2, 0.029177718832891247) (64.2, 0.02925531914893617) (65.2, 0.029411764705882353) (66.2, 0.029411764705882353) (67.2, 0.029490616621983913) (68.2, 0.02956989247311828) (69.2, 0.031578947368421054) (70.2, 0.03183023872679045) (71.2, 0.032) (72.2, 0.03208556149732621) (73.2, 0.03208556149732621) (74.2, 0.03208556149732621) (75.2, 0.032171581769437) (76.2, 0.03234501347708895) (77.2, 0.03234501347708895) (78.2, 0.034031413612565446) (80.2, 0.03439153439153439) (81.2, 0.034574468085106384) (82.2, 0.034574468085106384) (83.2, 0.034759358288770054) (84.2, 0.034759358288770054) (85.2, 0.035326086956521736) (86.2, 0.035422343324250684) (87.2, 0.03713527851458886) (88.2, 0.03753351206434316) (89.2, 0.04) (91.2, 0.04032258064516129) (92.2, 0.0427807486631016) (93.2, 0.04289544235924933) (94.2, 0.04289544235924933) (95.2, 0.0431266846361186) (96.2, 0.04336043360433604) (97.2, 0.044386422976501305) (98.2, 0.04712041884816754) (99.2, 0.047493403693931395) (100.2, 0.050397877984084884) (101.2, 0.051630434782608696) (102.2, 0.053763440860215055) (103.2, 0.053763440860215055) (104.2, 0.05630026809651475) (105.2, 0.05737704918032787) (106.2, 0.057441253263707574) (107.2, 0.05774278215223097) (108.2, 0.059431524547803614) (109.2, 0.060526315789473685) (110.2, 0.06504065040650407) (111.2, 0.07027027027027027) (113.2, 0.07923497267759563) (114.2, 0.08672086720867209) (115.2, 0.13372093023255813) (116.2, 0.2576419213973799)
        };
        
        \addplot+[
            ybar,
            fill=orange!50
        ]coordinates {
            (2, 0.008021390374331552) (4, 0.00804289544235925) (5, 0.00804289544235925) (12, 0.0106951871657754) (13, 0.0106951871657754) (20, 0.013404825737265416) (22, 0.015957446808510637) (30, 0.018666666666666668) (31, 0.01871657754010695) (38, 0.0213903743315508) (51, 0.024193548387096774) (53, 0.024324324324324326) (78, 0.03430079155672823) (89, 0.040106951871657755) (111, 0.07122507122507123)
        };
        \end{axis}
    \end{tikzpicture}
    \caption{Voice cloning Phoneme Error Rate (PER) of individual speakers from FormosaSpeech and Spontaneous Speech datasets.}
    \label{fig:vc}
\end{figure*}

\subsubsection{Comparing Code-Switching Abilities}

We compare the performance of various systems on code-switched terms of the TCCSD dataset across five prominently occurring categories to evaluate their suitability for modeling spontaneous bilingual speech. We assess the accuracy of each model using human annotations, focusing exclusively on the designated terms. A positive annotation is issued only when the pronunciations, including all lexical stress, are accurate. Evaluation results in Table~\ref{tab:code_comparison} show that BreezyVoice has great performance across all categories with English terms, and can even consistently maintain the correct flow in full English sentences while conditioning only on Chinese prompts. However, one minor limitation of the model lies in its handling of Chinese toponyms, a category that is less common in the English corpus. The improvement of this area is left as future work.

\begin{table}[h!]
\centering
\begin{tabular}{|l|c|c|c|c|c|}
\hline
\diagbox[width=2cm, height=1cm]{Cat.}{Model} & \makecell{\textbf{B} \\ \textbf{(ours)}}  & Z & Y&  U & M \\
\hline
\textbf{General}  & 8 & 5 & 8 & 8 & 7 \\
\hline
\textbf{Entities} & 9 & 6 & 4 & 7 & 4 \\
\hline
\textbf{Abbr.}    & 9 & 8 & 6 & 6 & 7 \\
\hline
\textbf{Toponyms}   & 3 & 3 & 7 & 3 & 4 \\
\hline
\textbf{Sentences} & 7 & 7 & 8 & 5 & 3 \\
\hline
\end{tabular}
\caption{Objective evaluation of BreezyVoice against four competing systems on the TCSSD dataset, measured by absolute accuracy scores.}
\label{tab:code_comparison}
\end{table}

\subsection{Evaluations in Voice Cloning}

We evaluate the model's voice cloning performance using our two curated speaker corpora comprising a total of 115 speakers. The performance across this diverse speaker profile provides a reasonable indication of how comprehensively the model represents the Taiwanese Mandarin-speaking population. From Figure~\ref{fig:vc}, it can be seen that performance on FormosaSpeech is generally better than Spontaneous Speech. Despite this, it is worth noting that over half of the spontaneous samples samples exhibit an error rate of less than 3\%. Excluding ASR inaccuracies, these samples are nearly error-free upon human inspection. Additionally, the speaker similarity averages 92.29\%, highlighting the high fidelity of the cloned voices. Generally speaking, voice cloning capabilities is robust across speakers.

\section{Analysis}
\label{sec:analysis}
At the extreme end of the spectrum of Figure~\ref{fig:vc}, some samples, likely from the long tail of the data distribution, exhibited catastrophic error rates of over 10 percent.  In this section, we look deeper into the failure modes of both speaker-cloning and word-pronunciation issues, aiming to uncover the underlying causes and identify potential mitigations.
\subsection{Model Robustness in Voice Cloning}
A manual review of these samples confirms that the audios are genuinely prone to errors. The most common issues include the random insertion of speech stopwords, followed by stuttering, and insertion of unintelligible hallucinated utterances. To address these errors, we carefully analyze the pipeline and evaluate the impact of each component in search of a potential resolution.
\subsubsection{Effects on CFM Conditions}
Working backwards, we begin by examining the CFM model. A brief recap of the original CFM formula is as follows:
\begin{equation}
\label{eq:standard}
    \textbf{X}_{output} = CFM(v, \textbf{U}_{cond}, \textbf{U}_{output}, \textbf{X}_{cond})
\end{equation}
In the case of an erroneous $\textbf{X}_{output}$, one or more elements from the set $\{v, \{\textbf{U}_{cond}, \textbf{X}_{cond}\}, \textbf{U}_{output}\}$ contribute to the error. To assess the impact of these factors, we designed two additional experiments, using $Z^{iconic}$ to denote an equivalent latent type of $Z$ generated from an top-line iconic speaker instead of the noisy speaker:
\begin{equation}
\label{eq:v_ablate}
    \textbf{X}^{I}_{output} = CFM(v^{Iconic}, \textbf{U}_{cond}, \textbf{U}_{output}, \textbf{X}_{cond})
\end{equation}
\begin{equation}
\label{eq:cond_ablate}
    \textbf{X}^{recon}_{cond} = CFM(v, \textbf{U}_{cond_b}, \textbf{U}_{cond}, \textbf{X}_{cond_b})
\end{equation}
Equation~\ref{eq:v_ablate} isolates the effect of the speaker embedding comparing with the standard process of Equation~\ref{eq:standard}, while Equation~\ref{eq:cond_ablate} unveils whether $\textbf{U}_{cond}$ is re-constructable from $\textbf{X}_{cond}$.
Our results show that $\textbf{X}_{output} \approx \textbf{X}^{I}_{output}$ and $\textbf{X}^{recon}_{cond} \approx \textbf{X}_{cond}$, which discounts the effect of $v$ and $\{\textbf{U}_{cond}, \textbf{X}_{cond}\}$ causing the failure.
These two results together point to $\textbf{U}_{output}$ predicted with the LLM being the primary source of error, backing claims made by \citet{du2024cosyvoice} that the speech units emcompass semantic content and prosody.
\subsubsection{Effects on LLM Conditions}

We now delve deeper into the causes of the noisy $\textbf{U}_{output}$. Recall the formula for generation of units with the LLM:
\begin{equation}
    \label{eq:1shot}
    \textbf{U}_{output} = LLM(v, \textbf{Y}_{cond}, \textbf{Y}_{output}, \textbf{U}_{cond})
\end{equation}
We performed experiments under the following dropout conditions:
\begin{equation}
    \label{eq:0shot}
    \textbf{U}^{0}_{output} = LLM(v, \textbf{Y}_{output})
\end{equation}
Our thorough benchmarking across all spontaneous speakers show that the zero-shot inference setting of Equation~\ref{eq:0shot} yield meaningfully different speech samples (units) than that of Equation~\ref{eq:1shot}. However, the evaluation metrics indicate that omitting the conditions does not result in more stable speech outputs: 61\% of speakers show deterioration while only 39\% of speakers improved, leading to an overall increase in PER. These results highlight the importance of an in-distribution LLM speaker embedding in generating coherent speech outputs.

\subsubsection{Iconic Units for Robust Speech Cloning}

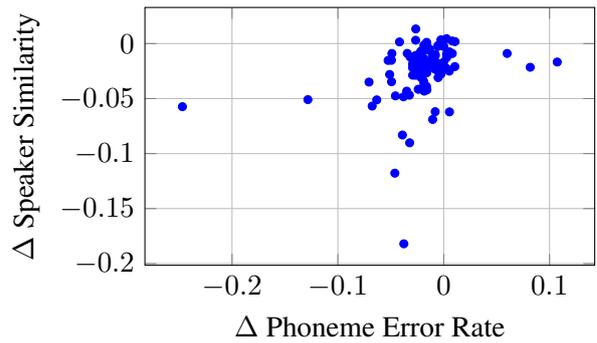
\begin{figure}[h!]
 \begin{tikzpicture}
        \begin{axis}[
            xlabel={$\Delta$ Phoneme Error Rate},
            ylabel={$\Delta$ Speaker Similarity},
            grid=major,
            width=7.5cm,  
            height=5cm,   
            legend pos=north east,
            ytick={0,-0.05,-0.1,-0.15,-0.2},
            scaled ticks=false,
            tick label style={/pgf/number format/fixed}
        ]
        \addplot[
            only marks,
            mark=*,
            mark size=1.5pt,
            color=blue,
        ] table {
            x   y
            -0.04617272587820219 -0.1177800000000001
            -0.12837333664967043 -0.050959999999999894
            -0.039024492681595135 -0.0831400000000001
            -0.008100550606820607 -0.02332000000000023
            0.005247311827956989 -0.025030000000000108
            -0.005426102333010097 -0.018039999999999945
            -0.03217158176943699 -0.09019999999999995
            -0.005290092576620092 -0.03092000000000006
            -0.010588652482269503 -0.019469999999999987
            -0.015622047244094488 -0.017350000000000088
            -0.018461930226636107 -0.027659999999999907
            -0.063277525869085 -0.051169999999999716
            -4.912073877591169e-05 -0.020510000000000028
            -0.01080157165349863 -0.011479999999999935
            -0.021093096160120287 -0.026739999999999986
            0.002773353054955057 -0.012829999999999897
            -0.049376737134675316 -0.015220000000000122
            0.08173971169495467 -0.02153000000000005
            0.10726817042606517 -0.016780000000000128
            -0.021461485948344523 -0.03079999999999994
            -0.0136860264519839 -0.005240000000000022
            -0.01597147950089127 -0.005749999999999922
            0.005448337825696317 -0.06211999999999995
            -0.005447493012378074 -0.00256999999999985
            -0.02406417112299465 -0.0415500000000002
            -0.02666666666666667 -0.02379000000000009
            -0.06759647347882643 -0.056820000000000204
            -0.01598573975044563 -0.026489999999999903
            -0.05108964406877121 -0.02811000000000008
            0.05990427667129844 -0.008970000000000034
            -0.01844543582704187 -0.04322999999999999
            -0.010922977364689484 -0.018429999999999946
            0.008107641883732964 -0.009009999999999962
            -0.0324314974703432 -0.04691999999999996
            -0.023336226446790044 -0.023329999999999962
            0.00011439029970258634 -0.0026800000000001267
            -0.007964043526257687 -0.011200000000000099
            -5.6889293434975174e-05 -0.01689000000000007
            0.002680965147453083 0.004429999999999823
            -0.016085790884718495 -0.017120000000000024
            -0.010695035460992907 -0.013340000000000019
            -0.0007556766575840188 -0.022419999999999995
            0.008100089365504914 0.0023300000000000542
            -0.029454582144195565 -0.018529999999999824
            -0.03178768493170116 -0.012089999999999823
            -0.015843056508808052 -0.04212000000000027
            -0.02214027445314241 -0.015940000000000065
            -0.008028520499108734 -0.06191000000000013
            -0.0026169074980088736 -0.027759999999999896
            -0.010666666666666668 -0.018199999999999772
            -0.024006670116727042 -0.010440000000000227
            -0.04895663799116485 -0.008989999999999831
            -0.07059183495060757 -0.03499000000000019
            -0.04931253305129561 -0.034680000000000044
            0.002680965147453083 -0.009529999999999927
            -0.021107432872138752 -0.021950000000000025
            -0.021842810780301963 -0.009220000000000228
            -0.0026096256684491996 0.003409999999999802
            0.01055296393218796 -0.020930000000000115
            -0.018822962027830185 -0.02355000000000007
            -0.021678346810422286 -0.007380000000000053
            -0.010404698072258392 -0.06899
            -0.026503922041447397 -0.02878999999999987
            -0.013114336643748408 -0.017700000000000382
            -0.03497498706227359 -0.04324000000000017
            -0.02648264514390275 0.003030000000000199
            -0.005375886524822696 -0.02475000000000005
            -0.013469610718187568 -0.004810000000000203
            -0.005190804859962352 -0.0016899999999999693
            -0.029339888448047836 -0.028739999999999988
            -0.016114387846291333 0.0011400000000000299
            -0.01886021505376344 -0.03449000000000024
            -0.02639153439153439 0.013289999999999913
            -0.03767764997549655 -0.18201000000000012
            -0.05250492541514214 -0.01529999999999998
            0.005404482876322678 -0.007530000000000259
            -0.01876675603217158 -0.041610000000000036
            -0.01566282015198424 -0.022439999999999682
            -0.03798380336817895 -0.048379999999999646
            -0.007992869875222816 -0.009769999999999945
            -0.018623771224307417 -0.0014799999999999258
            -0.03444043117557424 -0.008919999999999817
            -0.021526997030759033 -0.03022999999999998
            -0.24691806080756756 -0.05737000000000003
            -0.018503808564660492 -0.0038399999999999546
            -0.018659536541889482 -0.03739999999999988
            -0.019240296071803237 -0.01560999999999979
            -0.0189760295198697 -0.013449999999999851
            -0.010497128144186966 -0.010579999999999812
            -0.01865934872171116 -0.016670000000000185
            0.010695187165775402 0.0017899999999997362
            -0.016256462406041772 -0.03891
            -0.021361700907513874 -0.030399999999999983
            -0.02123324396782842 -0.006739999999999857
            -0.026216807949250544 -0.018349999999999866
            0.0026021505376344085 -0.022609999999999908
            -0.026349949910403115 -0.01071999999999984
            -0.041787085514834205 0.0014899999999999913
            -0.045605080930739346 -0.04752999999999996
            -0.028912280701754386 -0.014600000000000057
            -0.029411764705882356 -0.021909999999999985
        };
        \end{axis}
    \end{tikzpicture}
\caption{Speaker similarity sensitivity on phoneme error rate reduction, following the application of Iconic Unit Augmented Speech Cloning, in comparison to the standard speech cloning pipeline.}
\label{fig:corr}
\end{figure}

Given the importance of LLM speaker embeddings, we aim to explore the feasibility of a two-stage approach: first, consolidating content information through unit generations from an iconic speaker, and subsequently applying voice conversion to another (noisy) speaker using the CFM model. We dub this as \textbf{Iconic Unit Augmented Speech Cloning}. Our benchmark results show that this approach drastically reduces phoneme error rate from 3.4\% to 1.3\%, which is an overall reduction of 61.2\%, accompanied by only a slight trade-off in speaker similarity (-2.51\%). The reduction is evenly spread among speakers, where samples from 86 out of 100 speakers showed improvements. Figure~\ref{fig:corr} reveals a weak to moderate positive correlation of $r=0.29$ between these changes in error rate and speaker similarity, suggesting that voice cloning performance remains more constrained for samples prone to generating noisy units, despite the mitigative effects of the proposed approach.

\subsection{Model Robustness in Phonetic Identification}

While most pronunciations of most Chinese characters can be inherently derived from the LLM, as demonstrated by general benchmarking results, errors occur more frequently in challenging scenarios such as polyphone disambiguation and rare word identification. Our testing results on 23 hard instances requiring polyphone disambiguation show that the original pipeline failed in 8 cases, while g2pW augmentation successfully corrected all but 1 of the errors. Similar improvements were observed for rare words with fewer than 500 occurrences in the training corpus, where the error-inducing characters were effectively patched. Finally, we experimented with attaching incorrect Mandarin Phonetic Symbols to monophonic characters and found that the model adhered to the phonetic symbols rather than the correct character pronunciations, demonstrating a high level of phonetic attentiveness.

\section{Conclusion}

In this paper, we present BreezyVoice, a TTS system designed for Taiwanese Mandarin with enhanced polyphone disambiguation capabilities, which outperforms existing commercial TTS systems in both general and code-switching contexts. Experimental results show human parity cloning quality of BreezyVoice, which is further enhanced by our innovative iconic unit augmented speech cloning pipeline and phonetic augmentation techniques.

\bibliography{custom}

\appendix

\section{Appendix}
\label{sec:appendix}
\subsection{Subjective annotation instructions}
\label{sec:appendix_annotations}
In subjective evaluations, the annotators were tasked to evaluate the audio sample based on authenticity and audio quality
\begin{itemize}
\item Authenticity: Determine how authentic the voice sounds. Assess them based on prosody and intonation closeness to natural human speech.
\item Audio Quality: Determine the audio quality of the utterance. Assess it based on similarity to human articulation, and distinctiveness from a synthetic speech sample.
\end{itemize}

\section{Ethics Statement}

As BreezyVoice enables few-shot voice cloning; there is the inherent risk of potential misuse, such as voice spoofing. To ensure ethical use in real-world applications involving unknown speakers, it is essential to implement protocols that secure the speaker’s consent before utilizing their voice. Additionally, we are committed to developing anti-spoofing solutions to detect machine-generated speech, including but not limited to BreezyVoice, to alleviate these risks.
\end{document}